\documentclass[10pt,twocolumn,letterpaper]{article}

\usepackage{wacv}
\usepackage{times}
\usepackage{epsfig}
\usepackage{graphicx}
\usepackage{amsmath}
\usepackage{amssymb}
\usepackage[table]{xcolor}
\usepackage{tabularx,colortbl}
\usepackage[american]{babel}
\usepackage{multirow}
\usepackage{makecell}
\usepackage{capt-of,etoolbox}
\usepackage[accsupp]{axessibility}  % Improves PDF readability for those with disabilities.

\def\wacvPaperID{188} % Enter the WACV Paper ID here

\wacvfinalcopy % *** Uncomment this line for the final submission

\ifwacvfinal
\def\assignedStartPage{1} % *** Enter the assigned starting page number (instead of 9876)
\fi

\ifwacvfinal
\usepackage[breaklinks=true,bookmarks=false]{hyperref}
\else
\usepackage[pagebackref=true,breaklinks=true,colorlinks,bookmarks=false]{hyperref}
\fi

\ifwacvfinal
\else
\fi

\begin{document}

%%%%%%%%% TITLE
\title{Enhanced Bi-directional Motion Estimation for Video Frame Interpolation}

% \and

\author{Xin Jin$^1$ ~~~Longhai Wu$^1$ ~~~Guotao Shen$^1$~~~Youxin Chen$^1$ ~~Jie
    Chen$^1$ ~~~Jayoon Koo$^2$ ~~~Cheul-hee Hahm$^2$ \\ $^1$Samsung Electronics
    (China) R\&D Center ~~~~~~~~~$^2$Samsung Electronics, South Korea \\
    {\tt\small
    \{xin.jin,longhai.wu,guotao.shen,yx113.chen,ada.chen,j.goo,chhahm\}@samsung.com}
}

\twocolumn[{
\maketitle
\begin{center}
    \includegraphics[width=1.0\textwidth]{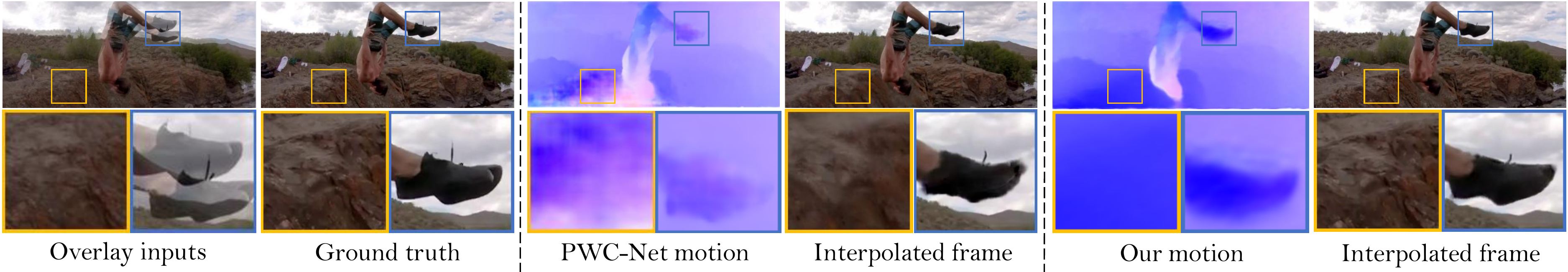}
    \captionof{figure}{\textbf{First two columns:} Overlay inputs and ground
        truth frame. \textbf{Middle two columns:} Motion field
        (from first to second frame) by PWC-Net~\cite{sun2018pwc} and
        corresponding interpolation. PWC-Net is end-to-end trained
    with our frame synthesis network. \textbf{Last two columns:} Motion field
and interpolated frame by our bi-directional motion estimator (15x smaller than
PWC-Net) and synthesis network.}
\label{fig:intro}
\end{center}
}]

%**************************************************************************************************%
% => Abastract
%**************************************************************************************************%
%%%%%%%%% ABSTRACT
\begin{abstract}

    We propose a simple yet effective algorithm for motion-based video frame
    interpolation. Existing motion-based interpolation methods typically rely on
    an off-the-shelf optical flow model or a U-Net based pyramid network for
    motion estimation, which either suffer from large model size or limited
    capacity in handling various challenging motion cases. In this work, we
    present a novel compact model to simultaneously estimate the bi-directional
    motions between input frames. It is designed by carefully adapting the
    ingredients (\eg, warping, correlation) in optical flow research for
    simultaneous bi-directional motion estimation within a flexible pyramid
    recurrent framework. Our motion estimator is extremely lightweight (15x
    smaller than PWC-Net), yet enables reliable handling of large and complex
    motion cases. Based on estimated bi-directional motions, we employ a
    synthesis network to fuse forward-warped representations and predict the
    intermediate frame. Our method achieves excellent performance on a broad
    range of frame interpolation benchmarks. Code and trained models are
    available at \url{https://github.com/srcn-ivl/EBME}.

\end{abstract}

%**************************************************************************************************%
% => Introduction
%**************************************************************************************************%
\section{Introduction}

Video frame interpolation aims to increase the frame rate of videos, by
synthesizing non-existent intermediate frames between original successive
frames. Increasing frame rate is beneficial for human
perception~\cite{kuroki2014effects}, and has wide applications in novel
view synthesis~\cite{flynn2016deepstereo}, video compression~\cite{lu2017novel},
adaptive streaming~\cite{wu2015modeling}, \etc.

The key challenge for frame interpolation is the possible complex, large motions
between input frames and intermediate frame. Based on whether a motion model is
employed to capture the per-pixel motion (\ie, optical flow) between frames,
existing methods can be classified into two categories: motion-agnostic
methods~\cite{niklaus2017video,meyer2018phasenet,cheng2020video,choi2020channel},
and motion-based
methods~\cite{jiang2018super,liu2017video,niklaus2018context,bao2019memc,niklaus2020softmax,park2020bmbc,park2021asymmetric,lu2022video}.
With recent advances in optical
flow~\cite{ilg2017flownet,hui2018liteflownet,sun2018pwc,teed2020raft},
motion-based interpolation has developed into a promising framework.

Motion-based interpolation involves two steps: (\romannumeral1) motion
estimation, and (\romannumeral2) frame synthesis. Motion field is estimated to
guide the synthesis of intermediate frame, by
forward-warping~\cite{niklaus2018context,niklaus2020softmax} or backward-warping
~\cite{jiang2018super,park2021asymmetric,sim2021xvfi} input frames towards
intermediate frame. Forward-warping is guided by motion from input frames to
intermediate frame, while backward-warping requires motion in reversed
direction. In particular, when the bi-directional motions between input frames
have been estimated, the motions from input frames to \textit{arbitrary}
intermediate frame required by forward-warping, can be easily approximated by
linearly scaling the motion magnitude
~\cite{niklaus2018context,niklaus2020softmax}.

Bi-directional motion estimation is a crucial step for most motion-based
interpolation
methods~\cite{niklaus2018context,niklaus2020softmax,jiang2018super,bao2019depth,sim2021xvfi}.
Many of existing
methods~\cite{niklaus2018context,bao2019depth,niklaus2020softmax} employ an
off-the-shelf optical flow model (\eg, PWC-Net~\cite{sun2018pwc}) for
bi-directional motions, which however suffer from large model size, need to run
the model twice, and can hardly handle extreme large motion beyond the training
data. Recently, a BiOF-I module~\cite{sim2021xvfi} is proposed for simultaneous
bi-directional motion estimation. It is based on a flexible pyramid recurrent
structure, which enables customizable pyramid levels in testing to handle large
motions.  At each pyramid level, BiOF-I uses current motion estimate to
backward-warp the features of both input frames towards each other, and employs
a shared plain U-Net to refine current motion.  However, U-Net is
over-simplified for optical flow, due to the lack of correlation volume, which
is a vital ingredient in modern optical flow
models~\cite{sun2018pwc,teed2020raft}.

% However, U-Net has limited capacity in
% handling complex motion cases like small fast-moving objects, due to the lack of
% correlation volume which is a vital ingredient in modern optical models
% ~\cite{sun2018pwc,teed2020raft}.
% Lee \etal~\cite{lee2022enhanced} integrated
% correlation into a pyramid recurrent network, but can only estimate
% single-directional motion at a time.

\begin{figure}[tb]
\centering
\includegraphics[width=0.48\textwidth]{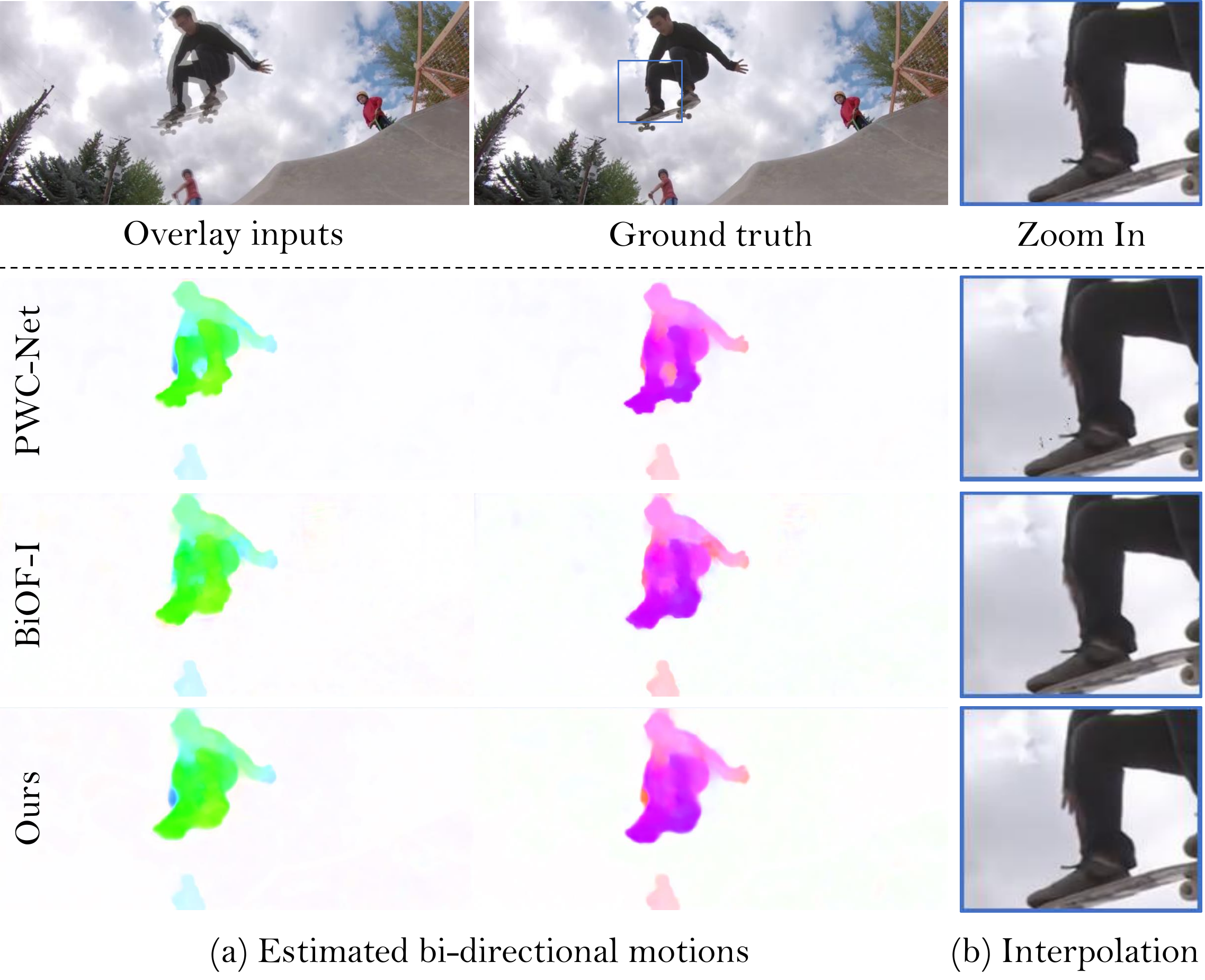}
\caption{Visual comparisons between PWC-Net~\cite{sun2018pwc},
BiOF-I~\cite{sim2021xvfi}, and our motion estimator, when combined with our
synthesis network for frame interpolation. BiOF-I fails to capture the motion of
fingers, due to the lack of correlation volume.}
\label{fig:motion-map}
\end{figure}

% \begin{figure}[tb]
% \centering
% \includegraphics[width=0.48\textwidth]{./figures/motion2.pdf}
% \caption{Visual comparisons between PWC-Net~\cite{sun2018pwc},
% BiOF-I~\cite{sim2021xvfi}, and our motion estimator, when combined with our
% synthesis network for frame interpolation. BiOF-I fails to capture the motion of
% fingers, due to the lack of correlation volume.}
% \label{fig:motion-map}
% \end{figure}

In this work, we present a simple but effective algorithm for frame
interpolation. Our main contribution is a novel bi-directional motion estimator.
Cast in a flexible pyramid recurrent framework, we adapt the ingredients (\eg,
warping, correlation) in optical flow research to simultaneously estimate the
bi-directional motions between input frames. In particular, at each pyramid
level, we forward-warp both input frames towards a hidden middle frame. This
middle-oriented forward-warping improves robustness against large motion, and
allows us to construct a single correlation volume for simultaneous
bi-directional motion estimation.  Based estimated bi-directional motions, we
forward-warp input frames and their context features to intermediate frame, and
employ a synthesis network to predict the intermediate frame from warped
representations.

Our bi-directional motion estimator enables better interpolation performance
than its single-directional counterpart which needs to run twice. It is 15x
smaller than PWC-Net~\cite{sun2018pwc}, yet can better handle large motion cases
and produce better interpolation result (see Figure~\ref{fig:intro}).  Compared
to BiOF-I~\cite{sim2021xvfi}, our motion estimator can capture the motion of
fast-moving small objects, giving better interpolation for local details (see
Figure~\ref{fig:motion-map}).

% Coupling these components enhances the
% ability in handling various challenging motion cases, compared to U-Net based
% pyramid recurrent network~\cite{sim2021xvfi}. Compared with off-the-shelf
% optical flow models~\cite{sun2018pwc,teed2020raft}, our motion estimator is much
% smaller, yet can better handle extreme large motion cases.

We conduct extensive experiments to verify the effectiveness of our
interpolation method named EBME -- \textbf{E}nhanced \textbf{B}i-directional
\textbf{M}otion \textbf{E}stimation for frame interpolation. Despite its small
model size, EBME performs favorably against state-of-the-art methods on a broad
range of benchmarks, from low resolution UCF101~\cite{soomro2012ucf101},
Vimeo90K~\cite{xue2019video}, to moderate-resolution
SNU-FILM~\cite{choi2020channel} and extremely high-resolution
4K1000FPS~\cite{sim2021xvfi}.

%**************************************************************************************************%
% => Related work
%**************************************************************************************************%
\section{Related Work}

\paragraph{Optical flow and correlation volume.} Optical flow is a low-level
vision task that aims to estimate the per-pixel motion between successive
frames.
% It has witnessed great progresses in past few
% years~\cite{dosovitskiy2015flownet,ilg2017flownet,hui2018liteflownet,sun2018pwc,teed2020raft}.
Modern optical flow models~\cite{sun2018pwc,hur2019iterative,teed2020raft} follow similar design
philosophy: extract CNN features for both input frames, construct correlation
volume with CNN features, and update the flow field upon a pyramid
structure~\cite{sun2018pwc} or at fixed high resolution~\cite{teed2020raft}.

Correlation volume, which stores the matching scores between the pixels of two
frames, is a discriminative representation for optical flow.  Before
constructing correlation volume, backward-warping is typically employed to align
the second frame to the first frame to compensate for estimated motion.  With
the warping operation (and down-sampled features), a partial correlation volume
with limited matching range is sufficient for optical flow
estimation~\cite{sun2018pwc}.

\paragraph{Off-the-shelf flow models for frame interpolation.}
PWC-Net~\cite{sun2018pwc} and RAFT~\cite{teed2020raft} are two representative
modern optical flow models. In particular, PWC-Net has been widely adopted in
frame interpolation to estimate the bi-directional motions by running
twice~\cite{bao2019depth,niklaus2018context,niklaus2020softmax}.  PWC-Net builds
a 6-level feature pyramids to handle large motion. At each level, it uses
current motion estimate to backward-warp the feature of second frame to the
first frame, constructs a correlation volume with warped feature and the feature
of first frame, and then infers a refined motion from correlation-injected
representation.

Off-the-shelf optical flow models have two disadvantages when applied for frame
interpolation. First, they typically have a large number of parameters.  Second,
when end-to-end trained with a synthesis network for frame interpolation, they
are prone to overfit the motion magnitude of training data.  Our bi-directional
motion estimator borrows some designs from modern optical flow models, but is
much more lightweight, robust to large motion, and specially-optimized for
simultaneous bi-directional motion estimation.

\begin{figure*}[!htb]
\centering
\includegraphics[width=1.0\textwidth]{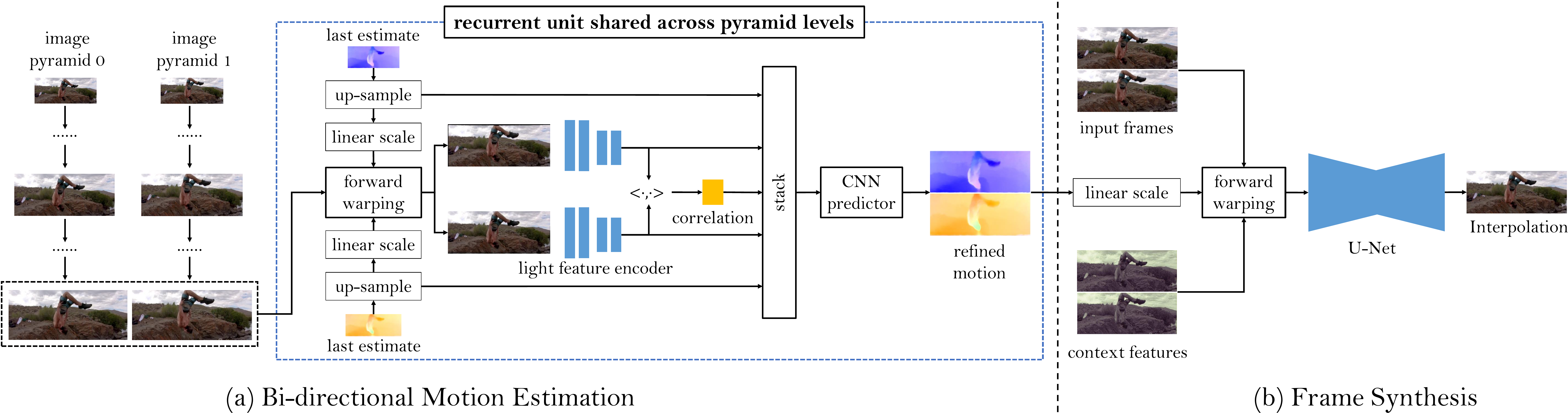}
\caption{
    Overview of our frame interpolation pipeline. \textbf{(a)} We repeatedly
    apply a novel recurrent unit across image pyramids to refine estimated
    bi-directional motions between input frames. The recurrent unit is
    integrated with middle-oriented forward-warping, lightweight feature
    encoder, and a single correlation volume for simultaneous bi-directional
    motion estimation. \textbf{(b)} Based on estimated bi-directional motions,
    we forward-warp input frames and their context features, and employ
    synthesis network to predict the intermediate frame.
}
\label{fig:pipeline}
\end{figure*}

\paragraph{U-Net motion estimator for frame interpolation.}
U-Net~\cite{ronneberger2015u} provides a powerful framework for dense prediction
tasks. In recent years, U-Net and U-Net based pyramid networks have been adopted
to estimate bi-directional motions~\cite{jiang2018super,sim2021xvfi} or
bilateral intermediate motions~\cite{zhang2020flexible,huang2020rife} for frame
interpolation.

However, due to the lack of correlation-based representations, these models
suffer from limited capacity in handling challenge motions (\eg, local complex
motion, small fast-moving objects). In addition, analogous to off-the-shelf
optical flow models, plain U-Net has difficulty in estimating extreme large
motion beyond the training data.

\paragraph{Flexible pyramid recurrent motion estimator.} With recurrent design
for both feature encoder and motion updater, recently proposed pyramid recurrent
motion estimators can flexibly handle extreme large motion
cases~\cite{zhang2020flexible,sim2021xvfi,lee2022enhanced}.  Since the recurrent
unit (base estimator) can be applied on pyramid structure for multiple times,
using a larger number of pyramid levels in testing can handle larger motions
beyond the training phase.

The BiOF-I module~\cite{sim2021xvfi} combines U-Net and pyramid recurrent
structure for simultaneous bi-directional motion estimation.  While BiOF-I
enables excellent high-resolution frame interpolation\footnote{This is achieved
    by training on 4K dataset, and combining extra module to approximate the
bilateral intermediate motions for backward-warping based frame synthesis.}, its
U-Net based recurrent unit is over-simplified to handle challenging motion
cases.  Lee \etal~\cite{lee2022enhanced} proposed Enhanced Correlation Matching
(ECM) within a pyramid recurrent network. However, it is not designed for
simultaneous bi-directional motion estimation. Furthermore, BiOF-I
backward-warps input frames towards each other and ECM forward-warps one input
frame towards another. Both warping strategies are not optimal in case of large
motions, based on our experiments.

\paragraph{Forward-warping for frame interpolation.} Compared to
backward-warping, the motion field required by forward-warping is easier to
acquire, and thus enables simpler pipeline for frame interpolation. However,
forward-warping is less adopted for frame interpolation, partially because it
may lead to holes in warped output.  Niklaus and Liu~\cite{niklaus2018context}
demonstrated that this issue may be remedied by warping both input frames. The
holes in one warped frame can be filled by the context information from anther
warped frame.  Another limitation of forward-warping is that multiple pixels in
source image may be mapped to the same target location. To solve this, softmax
splatting~\cite{niklaus2020softmax} is developed to adaptively assigns weights
to conflicted pixels.

With recent advances in forward-warping, we employ forward-warping for both
motion estimation and frame synthesis. In particular, we use the average
splatting operation in~\cite{niklaus2020softmax} as forward-warping, which
directly averages the conflicted pixels to generate the pixel in target
position. Average splatting is simpler than softmax splatting operation which
relies on a confidence map.

%**************************************************************************************************%
% => Approach
%**************************************************************************************************%
\section{Our Approach}

\subsection{Overview of the Pipeline}\label{subsec:pipe-overview} As shown in
Figure~\ref{fig:pipeline}, our frame interpolation pipeline involves two steps:
(a) bi-directional motion estimation, and (b) frame synthesis. Our main
innovation is the bi-directional motion estimator.

Formally, given two input frames $I_0$ and $I_1$, our goal is to predict the
intermediate frame $I_t$ at arbitrary time $t \in (0, 1)$. Firstly, we employ
our novel bi-directional motion estimator to calculate the motion $F_{0
\rightarrow 1}$ and $F_{1 \rightarrow 0}$ between $I_0$ and $I_1$, and linearly
scale them to obtain $F_{0 \rightarrow t}$ and $F_{1 \rightarrow t}$, \ie, the
motion from $I_0$ and $I_1$ to $I_t$:
\begin{equation}
    \begin{split}
    & F_{0 \rightarrow t} = t \cdot F_{0 \rightarrow 1} \\
    & F_{1 \rightarrow t} = (1-t) \cdot F_{1 \rightarrow 0}
    \end{split}
    \label{eq:linear-scale}
\end{equation}
With $F_{0 \rightarrow t}$ and $F_{1 \rightarrow t}$, we forward-warp input
frames and their context features, and feed warped representations into a
synthesis network to predict $I_t$. The synthesis network outputs a mask $M$ for
combining the warped frames, and a residual image $\Delta I_t$ for further
refinement.
\begin{equation}
    \small
    \begin{split}
    I_t \!=\! M \! \odot \! \mathop{\mathcal{W}} \limits ^{\rightarrow} (I_0, F_{0 \rightarrow t}) + \
    (1 \!-\! M) \odot \mathop{\mathcal{W}} \limits ^{\rightarrow} (I_1, F_{1 \rightarrow t}) + \Delta I_t\\
    \end{split}
    \label{eq:interp}
\end{equation}
where $\odot$ denotes element-wise multiplication, $\mathop{\mathcal{W}} \limits
^{\rightarrow}$ denotes the forward-warping operation (average
splatting~\cite{niklaus2020softmax}).

In testing, our bi-directional motion estimator can operate on flexible
customizable image pyramids to handle large motion.  Since motion magnitude
scales with resolution, we suggest a simple method to calculate the number of
pyramid levels in testing. Assume that the number of pyramid levels in training
is $L^{train}$, and the averaged width (or height) of test images is $n$ times
of training images. Then, we can calculate the number of test pyramid levels as
follows.
\begin{equation}
    L^{test} = \texttt{ceil}(L^{train} + \mathtt{log}_{2} n)
    \label{eq:level}
\end{equation}
where $\texttt{ceil}()$ rounds up a float number to get an integer.

\subsection{Bi-directional Motion Estimation}\label{subsec:our-flow}

\paragraph{Pyramid recurrent framework and recurrent unit.} As shown in
Figure~\ref{fig:pipeline} (a), the macro structure of our bi-directional motion
estimator is a pyramid recurrent network. Given two input frames, we firstly
construct image pyramids for them, then repeatedly apply a novel recurrent unit
across the pyramid levels to refine estimated bi-directional motions from
coarse-to-fine.

At each pyramid level, we first up-sample the estimated bi-directional motions
from previous level as initial motion (zero initialization for the top level).
Based on scaled initial motion, we forward-warp both input frames to a hidden
middle frame. Then, we employ an extremely lightweight feature encoder to
extract CNN features for both warped frames.  Lastly, we construct a correlation
volume with CNN features of warped frames, and estimate the bi-directional
motions from correlation injected features.

In the following, we detail the three key components involved in our
recurrent unit: \textit{middle-oriented forward-warping},
\textit{extremely lightweight feature encoder}, and \textit{correlation based
bi-directional motion estimation}.

\begin{figure}[tb]
\centering
\includegraphics[width=0.48\textwidth]{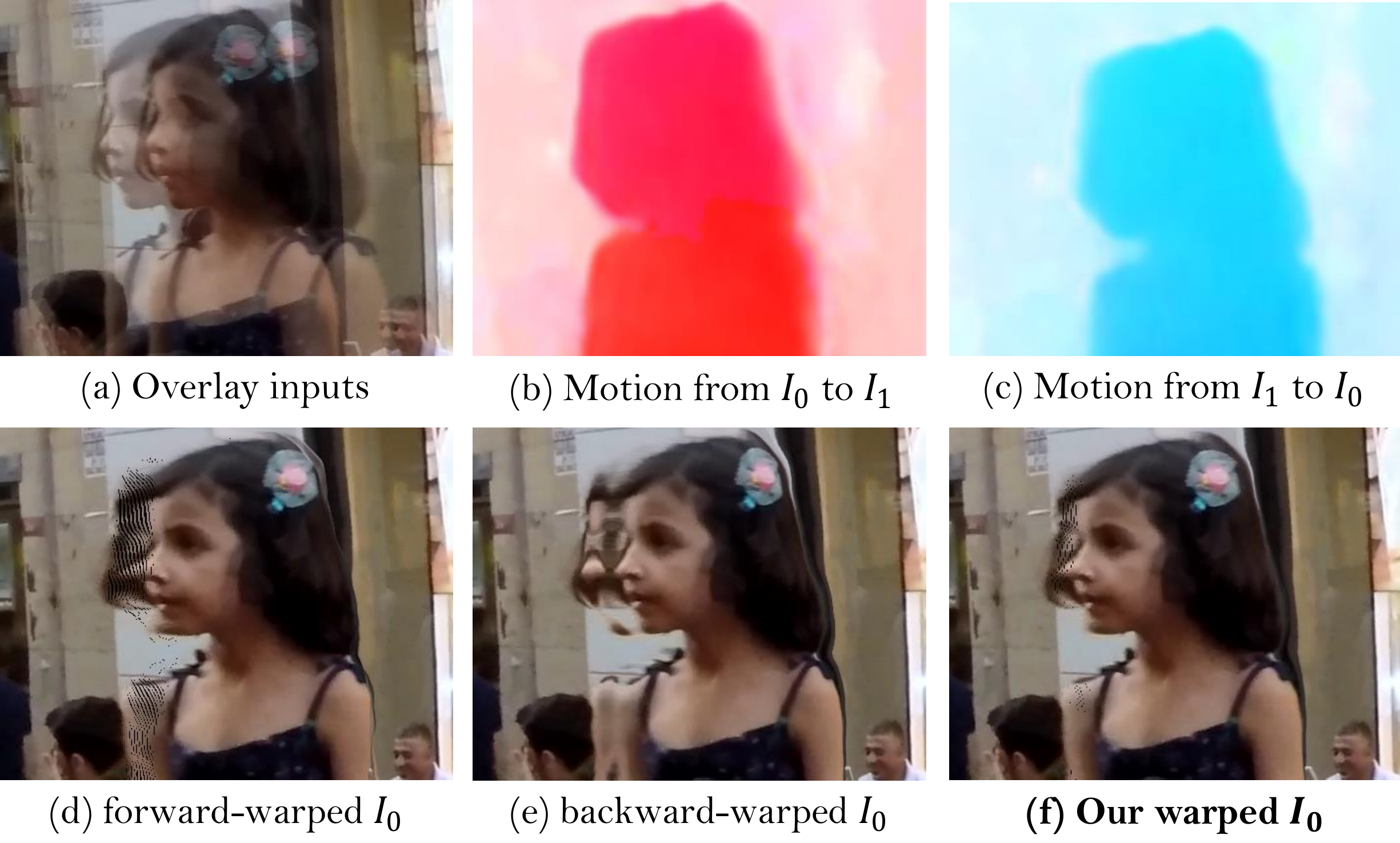}
\caption{Comparisons of different warping strategies in cases of large motion.
Our \textit{middle-oriented forward-warping} can reduce the possible artifacts
caused by warping, as it uses linearly-scaled motion that has smaller
magnitude.}
\label{fig:warp}
\end{figure}

\paragraph{Middle-oriented forward-warping.} Warping both input frames towards
each other is a natural idea for simultaneous bi-directional motion
estimation~\cite{sim2021xvfi}. However, this comes with two disadvantages.
First, it may lead to serious artifacts in warped output in case of large
motions (see Figure~\ref{fig:warp} (d) and (e)). Second, two (rather than one)
correlation volumes are required to record the matching scores between two
original frames and the frames warped towards them.

Considering these, at $i$-th pyramid level, we firstly forward-warp both input
frames $I_0^i$ and $I_1^i$ towards a hidden middle frame $I_{0.5}^i$, using
linearly-scaled motions that have smaller magnitude than initial motions. Due to
reduced motion magnitude, our middle-oriented forward-warping has the chance to
reduce the impacts of possible artifacts caused by warping (see
Figure~\ref{fig:warp} (f)).  Furthermore, warping both input frames to a hidden
frame allows us to construct a single correlation volume for simultaneous
bi-directional motion estimation.

\paragraph{Extremely lightweight feature encoder.} Pyramidal optical flow models
like PWC-Net~\cite{sun2018pwc} typically require a feature encoder with many
down-sampling layers to construct feature pyramids. To handle large motion,
PWC-Net employs a feature encoder of 6 down-sampling layers.

Our motion estimator handles large motion by customizing the number of pyramid
levels of \textit{outer} image pyramids. Thus, the feature encoder involved
in \textit{inner} recurrent unit does not need many down-sampling layers. We
employ an extremely lightweight feature encoder with only two down-sampling
layers to extract CNN features for both warped frames. It has
only about 0.1 M parameters, while PWC-Net's feature encoder has 1.7 M
parameters.

\paragraph{Correlation-based bi-directional motion estimation.} Existing works
construct a correlation volume between one original frame and another frame
warped towards it to estimate single-directional motion between input
frames~\cite{sun2018pwc,lee2022enhanced}. While for simultaneous bi-directional
motion estimation, two correlation volumes are required, if input frames are
warped towards each other.

Instead, we construct a \textit{single} correlation volume for simultaneous
bi-directional motion estimation, using CNN features of both warped frames that
have compensated for estimated bi-directional motions. Following
PWC-Net~\cite{sun2018pwc}, we set the local search range on the feature map of
the second warped frame as 4.  We concatenate the correlation volume, CNN
features, and up-sampled bi-directional motions to form input features, and
use a 6-layer convolutional network to predict the bi-directional motions. Since
our feature encoder has two down-sampling layers, the estimated motion is at 1/4
resolution of the input frame. We use bi-linear interpolation to up-scale the
motion to original scale.

\subsection{Frame Synthesis}\label{subsec:our-synthesis}

Based on estimated bi-directional motions, we employ a synthesis network to
predict the intermediate frame from forward-warped representations.

\paragraph{A simple baseline synthesis network.} Our synthesis network follows
the design of previous context-aware synthesis
networks~\cite{niklaus2020softmax,huang2020rife}, which take both warped frames
and warped context features as input. We extract 4-level pyramid context
features for both input frames.

We employ a simple U-Net as our synthesis network, which has four down-sampling
layers, and four up-sampling layers. It takes warped frames, warped context
features, original images, and bi-directional motions as input, and outputs a
mask $M$ for combining the warped frames, and a residual image $\Delta I_t$ for
further refinement (see Equation~\ref{eq:interp}). We refer to this synthesis
network as our \textit{base} synthesis network.

\paragraph{High-resolution synthesis with convex down-sampling.} Higher
resolution input often has advantages for dense prediction
tasks~\cite{redmon2018yolov3,liu2021swin}. We verify this for frame synthesis.
Specifically, we up-sample the input frames and estimated bi-directional motions
to 2x resolution, feed them to our synthesis network, and obtain a 2x resolution
interpolation. To recover the original scale, we add a lightweight head to our
synthesis network to predict $5\times5$ dynamic filters for the pixels with
stride 2 on the 2x resolution interpolation. These filters allow us to take a
convex weighted combination over $5\times5$ neighborhoods on the 2x resolution
interpolation to predict each pixel of the target frame of original scale.

This convex down-sampling strategy achieves better performance than bi-linear
down-sampling, 0.1 dB improvement on the ``extreme" subset of
SNU-FILM~\cite{choi2020channel}.  We refer to this structure as
\textit{high-resolution} synthesis network.

\subsection{Architecture Variants} We name our frame interpolation method as
EBME -- \textbf{E}nhanced \textbf{B}i-directional \textbf{M}otion
\textbf{E}stimation for frame interpolation. We construct three versions of
EBME, with almost the same model size but increased computational cost:
\begin{itemize}
    \item EBME: It combines our bi-directional motion estimator with the base
        version of synthesis network. 
    \item EBME-H: It combines our motion estimator with the high-resolution
        version of synthesis network.
    \item EBME-H$^{*}$: It uses the test-time augmentation (refer to
        Section~\ref{subsubsec:test-aug}) with EBME-H, which doubles the
        computational cost but further improves performance.
\end{itemize}

\subsection{Implementation Details}

\paragraph{Loss function.} For fair comparisons with recent works, all models
are trained only with the synthesis loss, without auxiliary supervision for
motion. Our loss is weighted sum of Charbonnier loss~\cite{charbonnier1994two}
and census loss~\cite{meister2018unflow} between ground truth $I_t^{GT}$ and our
interpolation $I_{t}$:
\begin{equation}
    L = \rho (I_t^{GT} - I_t) + \lambda \cdot L_{cen}(I_t^{GT}, I_t),
    \label{eq:loss}
\end{equation}
where $\rho(x) = (x^2 + \epsilon^2)^\alpha$ is the Charbonnier function,
$L_{cen}$ is the census loss, and $\lambda$ is a trade-off hyper-parameter. We
empirically set $\alpha=0.5$, $\epsilon=10^{-6}$, $\lambda = 0.1$.

\paragraph{Training dataset.} We train our model on the Vimeo90K
dataset~\cite{xue2019video}. Vimeo90K contains 51,312 triplets with resolution
of $448\times256$ for training. We augment the training images by randomly
cropping $256\times256$ patches. We also apply random flipping, rotating,
reversing the order of the triplets for data augmentation.

\paragraph{Optimization.} Our optimizer is AdamW~\cite{loshchilov2017decoupled}
with weight decay $10^{-4}$ for 0.8 M iterations, using a batch size of 32. We
gradually reduce the learning rate during training from $2\times10^{-4}$ to
$2\times10^{-5}$ using cosine annealing.

\paragraph{Test-time augmentation.} \label{subsubsec:test-aug} We verify a
practice strategy described in~\cite{huang2020rife}. We flip the input frames
horizontally and vertically to get augmented test data, and use our model to
infer two results and reverse the flipping. A more robust prediction can be
obtained by averaging these two results.

%**************************************************************************************************%
% => Experiments
%**************************************************************************************************%
\section{Experiments}

\begin{table*}[tb]
\centering
% \scriptsize
\setlength{\tabcolsep}{0pt}
\begin{tabular*}{1.0\textwidth}{@{\extracolsep{\fill}}*{10}{lcccccccc}}
\hline
\multirow{2}{*}{methods} & \multirow{2}{*}{UCF101} &
\multirow{2}{*}{Vimeo90K} & \multicolumn{4}{c}{SNU-FILM} & parameters & runtime \\
\cline{4-7}
& &  &  easy & medium & hard & extreme & (millions) & (seconds)\\
\Xhline{2\arrayrulewidth}
% CyclicGen~\cite{liu2019deep}  & 35.11/0.968 & 32.09/0.949 & 37.72/0.984 & 32.47/0.955
%                               & 26.95/0.887 & 22.70/0.808  & 19.8    & 0.09   \\
DAIN~\cite{bao2019depth}  & 34.99/0.968 & 34.71/0.976 & 39.73/\textcolor{blue}{\underline{0.990}} & 35.46/0.978  
                          & 30.17/0.934  & 25.09/0.858 & 24.0    & 0.15   \\
CAIN~\cite{choi2020channel}  & 34.91/\textcolor{blue}{\underline{0.969}} & 34.65/0.973
                             & 39.89/\textcolor{blue}{\underline{0.990}} & 35.61/0.978  & 29.90/0.929  & 24.78/0.851 & 42.8    & 0.04   \\
SoftSplat~\cite{niklaus2020softmax}  & \textcolor{blue}{\underline{35.39}}/0.952 & 36.10/0.970
                                     & - & - & - & - & -  &  -  \\
AdaCoF~\cite{lee2020adacof} & 34.90/0.968 & 34.47/0.973 & 39.80/\textcolor{blue}{\underline{0.990}} & 35.05/0.975  
                            & 29.46/0.924   & 24.31/0.844  & 22.9   & \textcolor{blue}{\underline{0.03}}   \\
BMBC~\cite{park2020bmbc} & 35.15/\textcolor{blue}{\underline{0.969}} & 35.01/0.976 & 39.90/\textcolor{blue}{\underline{0.990
    }}                         & 35.31/0.977  & 29.33/0.927  & 23.92/0.843  & 11.0   & 0.82   \\
% RIFE-Large~\cite{huang2020rife} & \textcolor{red}{\textbf{35.41}}/\textcolor{red}{\textbf{0.970}}
%                                 & 36.12/\textcolor{blue}{\underline{0.980}}
%                                 & - & - & - & - & \textcolor{blue}{\underline{9.8}}   & -   \\
ABME~\cite{park2021asymmetric} & 35.38/\textcolor{red}{\textbf{0.970}} 
                               & \textcolor{blue}{\underline{36.18}}/\textcolor{red}{\textbf{0.981}} 
                               & 39.59/\textcolor{blue}{\underline{0.990}} & 35.77/\textcolor{blue}{\underline{0.979}}
                               & \textcolor{blue}{\underline{30.58}}/\textcolor{blue}{\underline{0.936}}
                               & \textcolor{red}{\textbf{25.42}}/\textcolor{red}{\textbf{0.864}}
                               & 18.1 & 0.28   \\
XVFI$_v$~\cite{sim2021xvfi} & 35.18/0.952 & 35.07/0.968 & 39.78/0.984 & 35.37/0.964
                            & 29.91/0.894 & 24.73/0.778  & 5.5   & 0.10   \\
ECM$_v$~\cite{lee2022enhanced} & 34.97/0.951 & 34.95/0.975 & - & -
                            & - & -  &
\textcolor{blue}{\underline{4.7}}   & -   \\
\hline
EBME (ours) & 35.30/\textcolor{blue}{\underline{0.969}} & 35.58/0.978 
            & 40.01/\textcolor{red}{\textbf{0.991}} & 35.80/\textcolor{blue}{\underline{0.979}}
            & 30.42/0.935  & 25.25/0.861 & \textcolor{red}{\textbf{3.9}}  & \textcolor{red}{\textbf{0.02}}   \\
EBME-H (ours) & 35.35/\textcolor{blue}{\underline{0.969}} 
              & 36.06/\textcolor{blue}{\underline{0.980}} 
              & \textcolor{blue}{\underline{40.20}}/\textcolor{red}{\textbf{0.991}}
              & \textcolor{blue}{\underline{36.00}}/\textcolor{red}{\textbf{0.980}}
              & 30.54/\textcolor{blue}{\underline{0.936}}  & 25.30/0.862 
              & \textcolor{red}{\textbf{3.9}}  & 0.04   \\
EBME-H$^*$ (ours) & \textcolor{red}{\textbf{35.41}}/\textcolor{red}{\textbf{0.970}} 
                  & \textcolor{red}{\textbf{36.19}}/\textcolor{red}{\textbf{0.981}}
                  & \textcolor{red}{\textbf{40.28}}/\textcolor{red}{\textbf{0.991}}
                  & \textcolor{red}{\textbf{36.07}}/\textcolor{red}{\textbf{0.980}}
                  & \textcolor{red}{\textbf{30.64}}/\textcolor{red}{\textbf{0.937}}
                  & \textcolor{blue}{\underline{25.40}}/\textcolor{blue}{\underline{0.863}}
                  & \textcolor{red}{\textbf{3.9}}  & 0.08   \\
\hline
\end{tabular*}
\caption{Qualitative (PSNR/SSIM) comparisons to state-of-the-art methods on
UCF101~\cite{soomro2012ucf101}, Vimeo90K~\cite{xue2019video} and
SNU-FILM~\cite{choi2020channel} benchmarks. \textcolor{red}{\textbf{RED}}: best
performance, \textcolor{blue}{\underline{BLUE}}: second best performance.}
\label{tab:sota}
\end{table*}

\begin{figure*}[tb]
\centering
\includegraphics[width=1.0\textwidth]{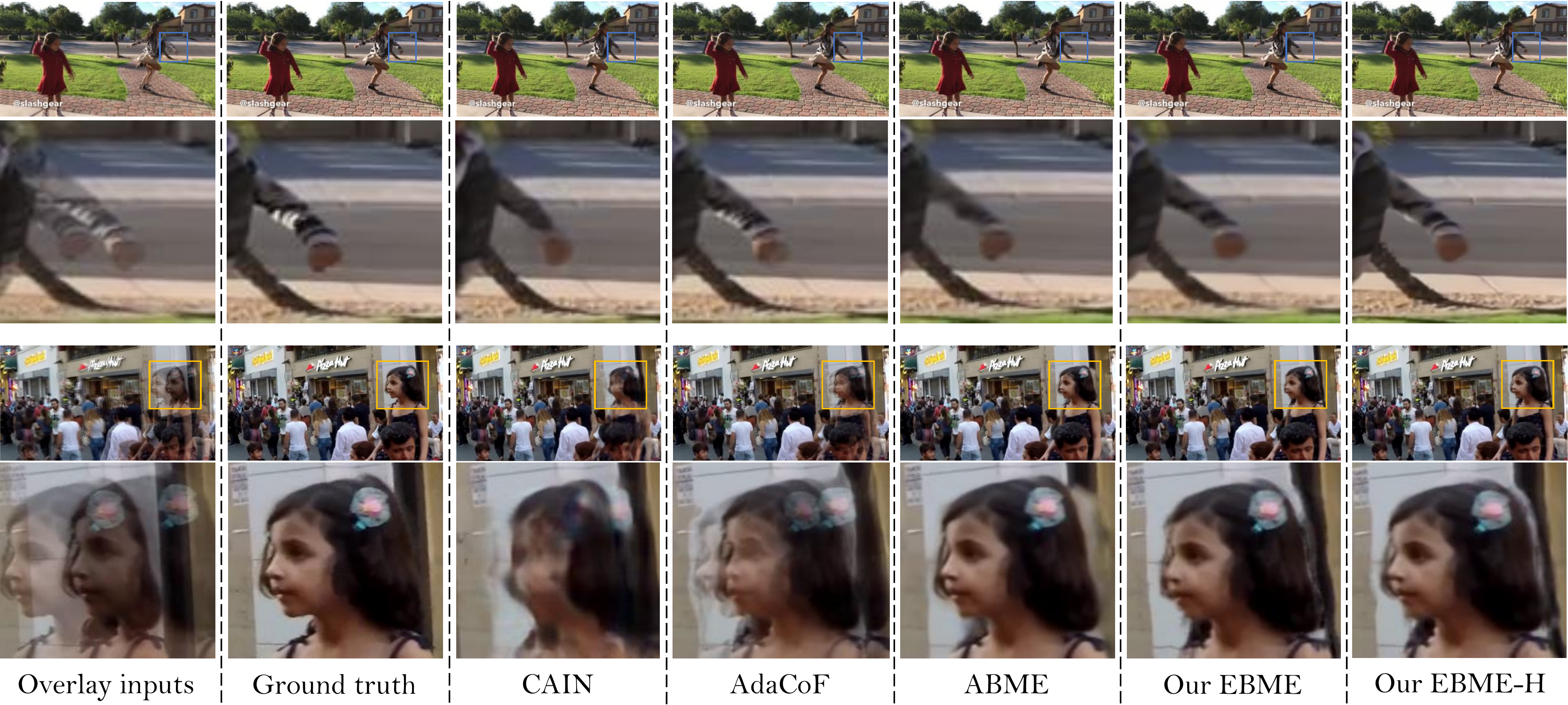}
\caption{Visual comparisons on two examples from the ``extreme" subset of
    SNU-FILM~\cite{choi2020channel}. The first two rows show the synthesis
    results for detailed textures, while the last two rows demonstrate the
    results with complex and
large motion.} \label{fig:snufilm}
\end{figure*}

\subsection{Experiment Settings}

\paragraph{Evaluation datasets.} While our method is trained only on
Vimeo90K~\cite{xue2019video}, we evaluate it on a broad range of benchmarks with
different resolutions.
\begin{itemize}
    \item \textbf{UCF101~\cite{soomro2012ucf101}}: The test set of UCF101
        contains 379 triplets with a resolution of 256$\times$256. UCF101
        contains a large variety of human actions.
    \item \textbf{Vimeo90K~\cite{xue2019video}}: The test set of Vimeo90K
        contains 3,782 triplets with a resolution of 448$\times$256.
    \item \textbf{SNU-FILM~\cite{choi2020channel}}: This dataset contains 1,240
        triplets, and most of them are of the resolution around 1280$\times$720.
        It contains four subsets with increasing motion scales -- easy, medium,
        hard, and extreme.
    \item \textbf{4K1000FPS~\cite{sim2021xvfi}}: This is a 4K resolution
        benchmark that supports multi-frame ($\times$8) interpolation.
\end{itemize}

\paragraph{Metrics.} We calculate peak signal-to-noise ratio (PSNR) and
structure similarity (SSIM) for quantitative evaluation of interpolation. For
the running time, we follow the practice of~\cite{park2021asymmetric}, and test
all models with a RTX 2080 Ti GPU for interpolating the ``Urban" sequence in
Middle-bury benchmark~\cite{baker2011database}, which has a resolution of
$640\times480$.

\paragraph{Customized number of pyramid levels.} We use 3-level image pyramids
when training on low-resolution Vimeo90K~\cite{xue2019video}. For benchmark
datasets, UCF101~\cite{soomro2012ucf101} has similar resolution with Vimeo90K,
SNU-FILM has a resolution of about 720p, and 4K1000FPS has a resolution of 4K.
Based on our suggested calculation method by Equation~\ref{eq:level}, we set the
test pyramid levels for UCF-101, SNU-FILM and 4K1000FPS as 3, 5 and 7,
respectively.

\begin{figure*}[tb]
\centering
\includegraphics[width=1.0\textwidth]{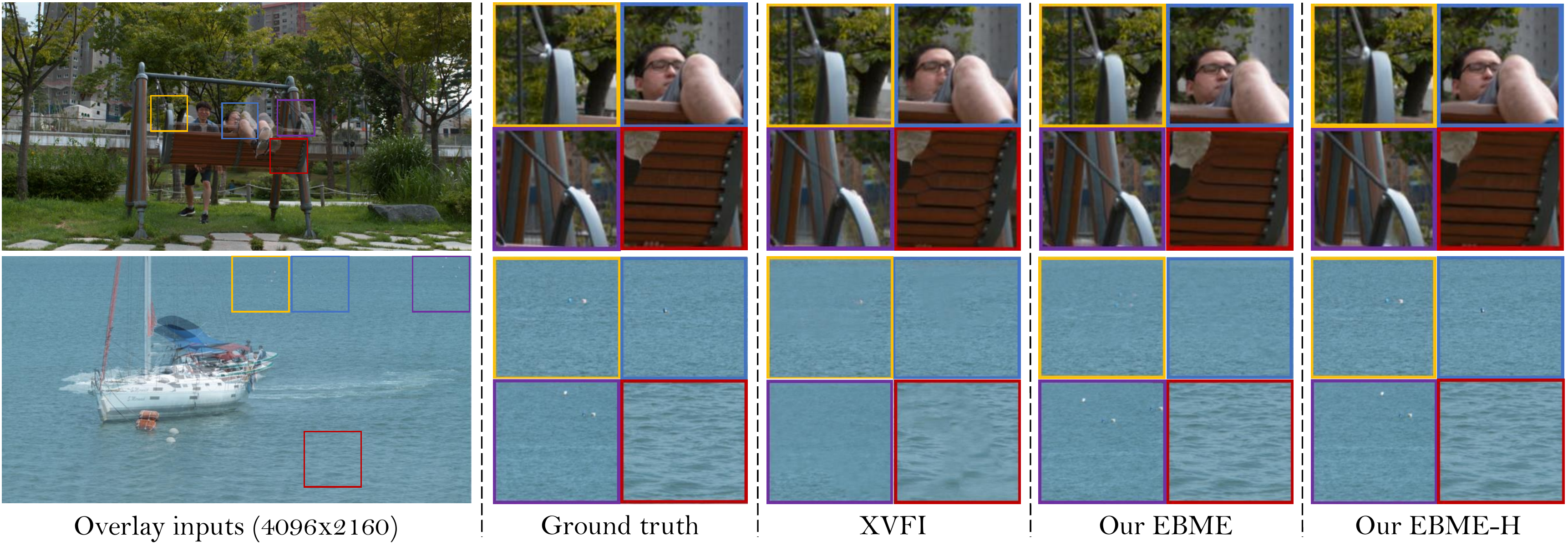}
\caption{Visual comparisons on 4K1000FPS~\cite{sim2021xvfi}.
XVFI~\cite{sim2021xvfi} trends to miss the moving small objects, while our
EBME-H gives interpolation results close to the ground truth.}
\label{fig:4k1000fps}
\end{figure*}

\subsection{Comparisons with State-of-the-art Methods}

We compare with state-of-the-art methods, including DAIN~\cite{bao2019depth},
CAIN~\cite{choi2020channel}, SoftSplat~\cite{niklaus2020softmax},
AdaCoF~\cite{lee2020adacof}, BMBC~\cite{park2020bmbc},
ABME~\cite{park2021asymmetric},  XVFI~\cite{sim2021xvfi}, and
ECM~\cite{lee2022enhanced}. We report their results by executing the source code
and trained models, except for SoftSplat and ECM which have not released the
full code. For SoftSplat and ECM, we copy the results from original paper.
To test XVFI$_v$ on SNU-FILM, we adjust the number of scale levels so that it
has the same down-sampling factor with our motion estimator.

\paragraph{Parameter and inference efficiency.} As shown the last two columns in
Table~\ref{tab:sota}, our frame interpolation algorithm has much less parameters
than state-of-the-art methods and runs very fast. In particular, due to the
macro recurrent design and the lightweight feature encoder, our bi-directional
motion estimator only has about 0.6 M parameters.

\paragraph{Low and moderate resolution frame interpolation.}
Table~\ref{tab:sota} reports the comparison results on low-resolution UCF101 and
Vimeo90K datasets. Our EBME-H$^*$ achieves best performance on both benchmarks.
Our EBME also outperforms many state-of-the-art models including DAIN, CAIN,
AdaCoF, BMBC, XVFI$_v$, and ECM.

Table~\ref{tab:sota} also reports the comparison results on SNU-FILM. Our EBME-H
and EBME-H$^*$ perform similar with ABME~\cite{park2021asymmetric} on the
``hard" and ``extreme" subsets, but have better performance on the ``easy" and
``medium" subsets. It is worth noting that our models are about 4.5x smaller
than ABME, and run much faster.

Figure~\ref{fig:snufilm} gives two examples from the ``extreme" subset from
SNU-FILM. Our methods produce better interpolation results than ABME for some
detailed textures (first two rows), and give promising results for large motion
cases (last two rows), much better than CAIN and AdaCoF, and sightly better than
ABME.

\paragraph{4K resolution multiple frame interpolation.}

\begin{table}[tb]
\centering
% \scriptsize
\setlength{\tabcolsep}{0pt}
\begin{tabular*}{0.48\textwidth}{@{\extracolsep{\fill}}*{6}{lcccc}}
\hline
\multirow{2}{*}{methods} & \multirow{2}{*}{arbitrary} & \multirow{2}{*}{reuse flow} & \multicolumn{2}{c}{4K1000FPS} \\
\cline{4-5}
                         & &  & ~~~~PSNR & SSIM \\
\Xhline{2\arrayrulewidth}
DAIN~\cite{bao2019depth}  & \checkmark & \checkmark &  ~~~~26.78   & 0.807   \\
AdaCoF~\cite{lee2020adacof} & $\times$ & $\times$  & ~~~~23.90   & 0.727  \\
ABME~\cite{park2021asymmetric} & \checkmark & $\times$  & ~~~~\textcolor{red}{\textbf{30.16}}  & 0.879 \\
XVFI~\cite{sim2021xvfi} & \checkmark & partial & ~~~~\textcolor{blue}{\underline{30.12}}  & 0.870  \\
\hline
EBME (ours) & \checkmark & \checkmark & ~~~~27.86  & 0.881 \\
EBME-H (ours) & \checkmark & \checkmark & ~~~~28.72 & \textcolor{blue}{\underline{0.889}}  \\
EBME-H$^*$ (ours) & \checkmark & \checkmark & ~~~~29.46 & \textcolor{red}{\textbf{0.902}}  \\
\hline
\end{tabular*}
\caption{Comparisons on 4K1000FPS~\cite{xue2019video} for 8x interpolation.}
\label{tab:multi-interp}
\end{table}

Table~\ref{tab:multi-interp} reports the 8x interpolation results on 4K1000FPS.
Our method achieves the best performance by SSIM, but slight inferior results to
ABME and XVFI by PSNR. Note that XVFI is trained on 4K high-resolution data,
while other models are trained on low-resolution data. Our method supports
arbitrary-time frame interpolation, and can fully re-use estimated
bi-directional motions when interpolating multiple intermediate frames at
different time positions. By contrast, while XVFI~\cite{sim2021xvfi} can reuse
the bi-directional motions, it must refine the approximated intermediate flow
with an extra network at each time position.

Figure~\ref{fig:4k1000fps} shows two interpolation examples. Our methods give
better performance for moving small objects. The U-Net based pyramid motion
estimator in XVFI might have difficulty in capturing the motion of extreme small
objects.

\subsection{Analysis of Our Motion Estimator} We present analysis of our motion
estimator on the ``hard" and ``extreme" subsets of
SNU-FILM~\cite{choi2020channel}, which contain various challenging motion cases.

\paragraph{Design Choices of Motion Estimator.} In
Table~\ref{tab:ablation_flow}, we report the ablation results for the design
choices of our bi-directional motions estimator.

\begin{table}[tb]
\centering
\setlength{\tabcolsep}{0pt}
\begin{tabular*}{0.48\textwidth}{@{\extracolsep{\fill}}*{5}{c}}
\hline
\multirow{2}{*}{experiments} & \multirow{2}{*}{methods} & \multicolumn{2}{c}{SNU-FILM (PSNR $\uparrow$)}\\
\cline{3-4}
                             &   & ~~~hard & extreme \\
\Xhline{2\arrayrulewidth}
\multirow{2}{*}{bi-directional}& \cellcolor{gray!40} simultaneous & ~~~30.42 & 25.25 \\
                              & single-direction & ~~~30.19 & 25.12 \\
\hline
\multirow{3}{*}{warping type} & forward & ~~~30.36 & 25.21 \\
                              & \cellcolor{gray!40}middle-forward & ~~~30.42 & 25.25 \\
                              & backward & ~~~30.28 & 25.11 \\
\hline
\multirow{3}{*}{feature encoder} & 1-stage & ~~~30.36 & 25.20 \\
                             & \cellcolor{gray!40}2-stage & ~~~30.42 & 25.25 \\
                                    & 3-stage & ~~~30.26 & 25.15 \\
\hline
\multirow{2}{*}{correlation}  
                              & without & ~~~30.29 & 25.17 \\
                              & \cellcolor{gray!40}with & ~~~30.42 & 25.25 \\
\hline
\multirow{4}{*}{test pyramid}     & 3-level  & ~~~30.15 & 24.80  \\
                                  & 4-level  & ~~~30.42 & 25.20 \\
                                  & \cellcolor{gray!40} 5-level & ~~~30.42 & 25.25 \\
                                  & 6-level & ~~~30.40 & 25.22 \\
\hline
% \end{tabularx}
\end{tabular*}
\caption{Impacts of the design choices of our bi-directional motion estimator,
integrated with base synthesis network for frame interpolation. Default settings
are marked in \colorbox{gray!40}{gray}.}
\label{tab:ablation_flow}
\end{table}

\begin{figure}[tb]
\centering
\includegraphics[width=0.48\textwidth]{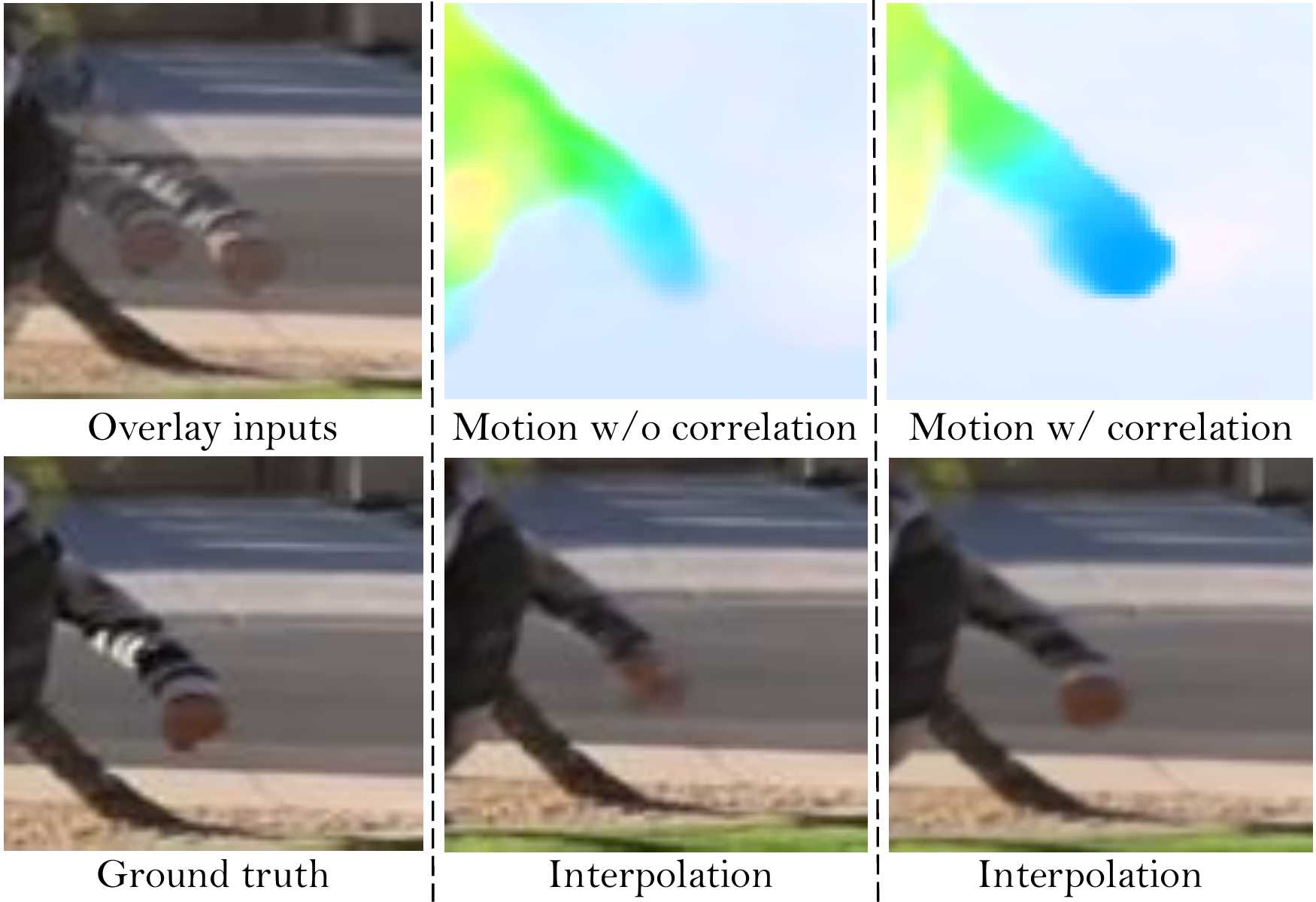}
\caption{Without correlation volume, our estimator may fail to estimate complex
motion, and lead to artifacts on interpolated frame.}
\label{fig:corr}
\end{figure}

\begin{itemize}
    \item \textbf{Simultaneous bi-directional estimation}: Our bi-directional
        motion estimator performs better than its single-directional counterpart
        that forward-warps the first frame to the second and constructs a
        correlation volume with warped frame and second frame.  We run the
        single-directional counterpart twice to obtain bi-directional motions.
        We verify that simultaneous bi-directional motion estimation can improve
        performance, and our middle-oriented warping also improves robustness
        against large motion.
    \item \textbf{Warping type}: Our middle-oriented forward-warping (denoted as
        ``middle-forward") achieves better performance than forward-warping and
        backward-warping that align input frames towards each other. Note that
        aligning input frames to each other needs to build two correlation
        volumes for the original two frames and warped frames, while our warping
        method enables the building of single correlation volume.
    \item \textbf{Feature encoder}: We investigate three settings for our
        feature encoder: one convolutional stage of 9 layers; two-stage with 3
        layers for first stage, and 6 layers for second stage; three-stage with
        3 layers for each stage. We double the number of filters with
        down-sampling layers. More down-sampling layers might be beneficial for
        large motion, but may lead to rough estimate. Two-stage feature encoder
        achieves the best trade-off.
    \item \textbf{Correlation volume}: Removing correlation volume from our
        motion model leads to inferior quantitative results.  Furthermore, as
        shown in Figure~\ref{fig:corr}, without a correlation
        volume, our estimator may have difficulty in estimating complex
        nonlinear motions, and lead to blurry artifacts in local regions.

        % Although these artifacts in local regions might have little impact on
        % quantitative metric, they may have bad affects for human perception.
    % \item \textbf{Weight sharing}: We compare our recurrent pyramid estimator
        % with a non-recurrent version that train three separate inner
        % bi-directional optical flow estimators. When both using a 3-level
        % image pyramid for testing, tying weights achieves better performance
        % with much less parameters.
    \item \textbf{Test pyramid level}: A 5-level image pyramid achieves good
        performance on the ``extreme" subset. Further increasing pyramid level
        does not lead to better results. This is consistent with our suggested
        calculation method described by Equation~\ref{eq:level}.
\end{itemize}

\paragraph{Motion Quality Comparison.}

\begin{table}[tb]
\centering
\setlength{\tabcolsep}{0pt}
\begin{tabular*}{0.48\textwidth}{@{\extracolsep{\fill}}*{6}{c}}
\hline
\multirow{2}{*}{experiments} & \multirow{2}{*}{methods}
                             & \multicolumn{2}{c}{SNU-FILM (PSNR $\uparrow$)}
                             & param.\\
\cline{3-4}
                             &   & ~~~hard & extreme & (M) \\
\Xhline{2\arrayrulewidth}
\multirow{3}{*}{warp approx.}
                             % & PWC (fix)  & ~~~27.30  & 23.09 & 9.4  \\
                             % & PWC (ft)  & ~~~28.29  & 23.81 & 9.4  \\
                             % & RAFT (rand)  & ~~~27.88  & 23.21 & 5.3  \\
                             & PWC-Net & ~~~\textcolor{blue}{\underline{28.37}}  & 23.59 & 9.4 \\
                             & BiOF-I & ~~~28.13  &
                             \textcolor{blue}{\underline{23.68}} & \textcolor{blue}{\underline{2.6}} \\
                             & Ours & ~~~\textcolor{red}{\textbf{28.62}} &  \textcolor{red}{\textbf{24.00}}
                             & \textcolor{red}{\textbf{0.6}} \\
\hline
\multirow{3}{*}{full pipeline} 
                             % & PWC (fix) & ~~~30.15 & 25.08 & 12.7  \\
                             %   & PWC (ft) & ~~~30.29 & 24.80 & 12.7 \\
                             % & RAFT (rand) & ~~~30.25 & 24.69 & 8.6  \\
                               & PWC-Net & ~~~\textcolor{blue}{\underline{30.04}} & 24.53 &12.7   \\
                               & BiOF-I & ~~~30.03  &
                             \textcolor{blue}{\underline{24.80}}  & \textcolor{blue}{\underline{5.9}} \\
                               & Ours & ~~~\textcolor{red}{\textbf{30.42}} &  \textcolor{red}{\textbf{25.25}}
                               & \textcolor{red}{\textbf{3.9}}\\
\hline
\end{tabular*}
\caption{Quantitative results of frame interpolation, enabled by
PWC-Net~\cite{sun2018pwc}, BiOF-I~\cite{sim2021xvfi}, and our motion estimator.}
\label{tab:ablation_motion}
\end{table}

We compare our bi-directional motion estimator with PWC-Net~\cite{sun2018pwc}
and BiOF-I~\cite{sim2021xvfi} for frame interpolation.  We end-to-end train
PWC-Net and BiOF-I from scratch with our basic synthesis network.  We
adjust the number of scale levels for BiOF-I so that it has the same
down-sampling factor with our bi-directional motion estimator when testing on
SNU-FILM.

We compare motion estimators for frame interpolation from two aspects:
interpolation by averaging two forward-warped frames, and interpolation by our
full pipeline. As shown in Table~\ref{tab:ablation_motion}, our motion estimator
enables much better interpolation results on the ``extreme" subset. In addition,
it is much smaller in size than PWC-Net and BiOF-I.

%**************************************************************************************************%
% => Conclusion
%**************************************************************************************************%

\section{Conclusion}

This work presented a lightweight yet effective frame interpolation algorithm,
based on a novel bi-directional motion estimator. Our method achieved
excellent performance on various frame interpolation benchmarks. This work
aims at motion-based frame interpolation, and does not pursue the motion
accuracy on optical flow benchmarks. In the future, we will verify the
effectiveness of our lightweight motion estimator for general-purpose optical
flow.

{\small
\bibliographystyle{ieee_fullname}
\bibliography{arxiv_paper}
}

\end{document}